\documentclass[conference]{IEEEtran}
\IEEEoverridecommandlockouts
\usepackage{cite}
\usepackage{amsmath,amssymb,amsfonts}
\usepackage{algorithmic}
\usepackage{graphicx}
\usepackage{textcomp}
\usepackage{xcolor}
\usepackage{comment}
\def\BibTeX{{\rm B\kern-.05em{\sc i\kern-.025em b}\kern-.08em
    T\kern-.1667em\lower.7ex\hbox{E}\kern-.125emX}}

\begin{document}

\title{Engineering AI Systems: A Research Agenda\\
\thanks{This research is in part supported by Software Center and by CHAIR (Chalmers AI Research Center).}
}

\author{
\IEEEauthorblockN{Jan Bosch}
\IEEEauthorblockA{\textit{Computer Science and Engineering} \\
\textit{Chalmers University of Technology}\\
Gothenburg, Sweden \\
jan.bosch@chalmers.se}
\and
\IEEEauthorblockN{Helena Holmström Olsson}
\IEEEauthorblockA{\textit{Computer Science and Media Technology} \\
\textit{Malmö University}\\
Malmö, Sweden \\
helena.holmstrom.olsson@mau.se}
\and
\IEEEauthorblockN{Ivica Crnkovic}
\IEEEauthorblockA{\textit{Computer Science and Engineering} \\
\textit{Chalmers University of Technology}\\
Gothenburg, Sweden \\
ivica.crnkovic@chalmers.se}
}

\maketitle

\begin{abstract}
Artificial intelligence (AI) and machine learning (ML) are increasingly broadly adopted in industry, However, based on well over a dozen case studies, we have learned that deploying industry-strength, production quality ML models in systems proves to be challenging. Companies experience challenges related to data quality, design methods and processes, performance of models as well as deployment and compliance. We learned that a new, structured engineering approach is required to construct and evolve systems that contain ML/DL components. In this paper, we provide a conceptualization of the typical evolution patterns that companies experience when employing ML as well as an overview of the key problems experienced by the companies that we have studied. The main contribution of the paper is a research agenda for AI engineering that provides an overview of the key engineering challenges surrounding ML solutions and an overview of open items that need to be addressed by the research community at large. 
\end{abstract}


\section{Introduction}
The prominence of artificial intelligence (AI) and specifically machine- and deep-learning (ML/DL) solutions has grown exponentially~\cite{amershi2019software, bernardi2019150}. Because of the Big Data era, more data is available than ever before and this data can be used for training ML/DL solutions. In parallel, progress in high-performance parallel hardware such as GPUs and FPGAs allows for training solutions of scales unfathomable even a decade ago. These two concurrent technology developments are at the heart of the rapid adoption of ML/DL solutions.

Virtually every company has an AI initiative ongoing and the number of experiments and prototypes in industry is phenomenal. Although earlier the province of large Software-as-a-Service (SaaS) companies, our research shows a democratization of AI and broad adoption across the entire industry, ranging from startups to large cyber-physical systems companies. ML solutions are deployed in telecommunications, healthcare, automotive, internet-of-things (IoT) as well as numerous other industries and we expect an exponential growth in the number of deployments across society.

Unfortunately, our research~\cite{arpteg2018software, lwakatare2019taxonomy, munappy2019data} shows that the transition from prototype to production-quality deployment of ML models proves to be challenging for many companies. Though not recognized by many, the engineering challenges surrounding ML prove to be significant. In our research, we have studied well over a dozen cases and identified the problems that these companies experience as they adopt ML. These problems are concerned with a range of topics including data quality, design methods and processes, performance of models as well as deployment and compliance. 

To the best of our knowledge, no papers exist that provide a systematic overview of the research challenges associated with the emerging field of AI engineering (which we define as an extension of Software Engineering with new processes and technologies needed for development and evolution of AI systems, i.e. systems that include AI components(. 
In this paper we provide a research agenda that has been derived from the research that we have conducted to date. The goal of this research agenda is to provide inspiration for the software engineering research community to start addressing the AI engineering challenges.

The purpose and contribution of this paper is threefold. First, we provide a conceptualization of the typical evolution patterns concerned with adoption of AI that companies experience. Second, we provide an overview of the engineering challenges surrounding ML solutions. Third, we provide a research agenda and overview of open items that need to be addressed by the research community at large.

The remainder of this paper is organized as follows. In the next section, we present the method underlying the research in this paper. In section~\ref{AIEchallenge}, we present on overview of the problems that we identified in our earlier research as well as a model capturing the evolution pattern of companies adopting AI solutions. Subsequently, we present our research agenda in section~\ref{Agenda}. Finally, we conclude the paper in section~\ref{Conclusion}.

\section{Research Method}
In the context of Software Center\footnote{https://www.software-center.se/}, we work with more than a dozen large international 
Cyber-physical systems (CPS) and embedded systems (ES)  companies, including Ericsson, Tetra Pak, Siemens, Bosch, Volvo Cars, Boeing and several others around, among other topics, the adoption of ML/DL technologies. In addition, we frequently have the opportunity to study and collaborate with companies also outside of Software Center that operate as SaaS companies in a variety of business domains.

Due to the limited length of this paper, we do not provide a full overview of all these case companies and our learning from each of these, nor the detailed research methods and activities that were employed in our various research activities. Instead, and for the purpose of this publication, we have selected a set of 16 primary cases as the foundation for the challenges we identify and the research agenda we outline. However, it should be noted that the work reported on in this paper is based also on learning from more than 20 companies from around the world, though with a focus on the software-intensive embedded systems industry in Europe, mostly Nordic countries. With this as our basis, we believe that the challenges we identify, and the research agenda we outline, reflect the key engineering challenges that companies in a variety of domains experience when employing and integrating ML/DL components in their systems. Below, we present the research approach adopted in this work and the cases we selected as the basis for this paper.

\subsection{Research approach and selected cases}

The goal of this research is to provide an understanding of the typical evolution patterns that companies experience, and the challenges they face, when adopting and integrating ML/DL components in their systems. Based on this understanding, we develop a research agenda in which we identify the open research questions that need to be addressed by the research community.

\begin{figure}
\includegraphics[width=0.5\textwidth]{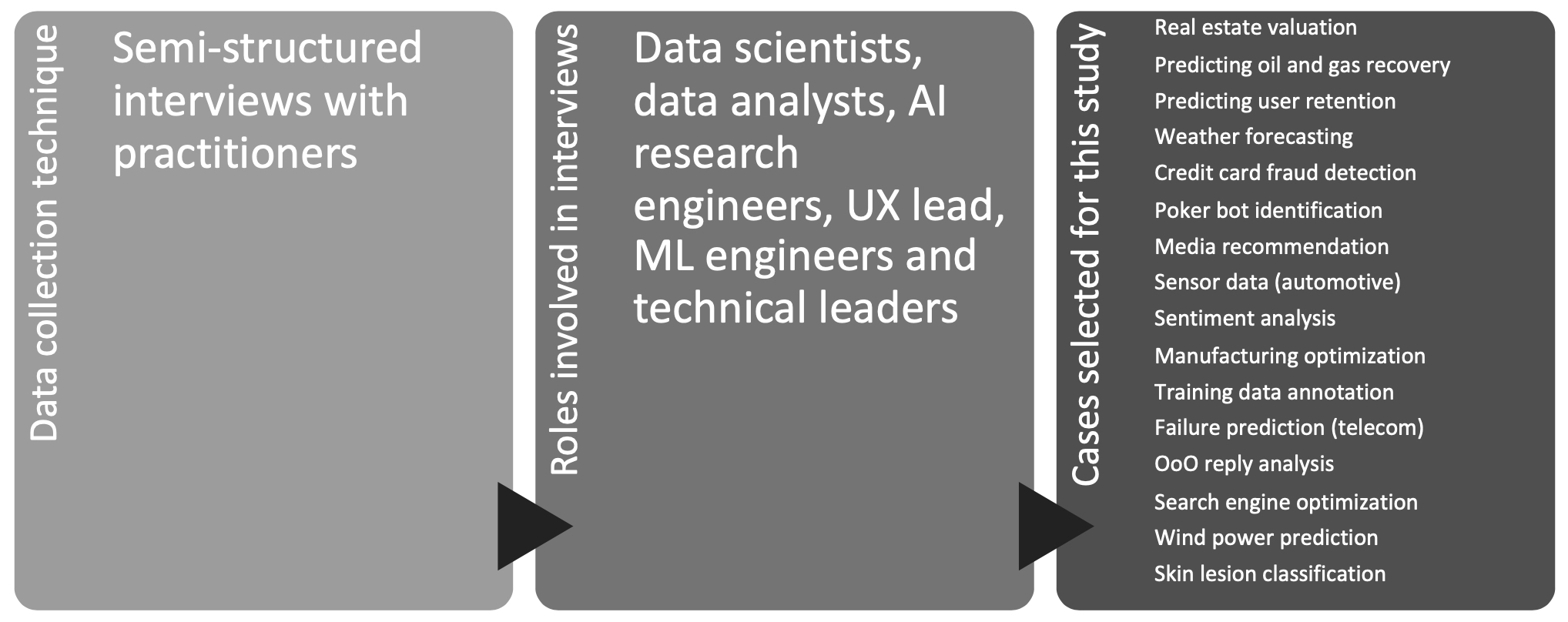}
\caption{Roles and cases that were selected as the basis for this study}
\label{RolesCases}
\end{figure}

In alignment with this research goal, our research builds on multiple-case study research~\cite{maxwell2012qualitative, flick2018designing} with semi-structured interviews and observations as the primary techniques for data collection. The findings we present in this paper build on a total number of 16 cases representing startups as well as large multinational companies in domains such as e.g. real estate, weather forecasting, fraud detection, sentiment analysis and failure prediction. Each case represents a software-intensive system that incorporates ML and DL components and involves challenges ranging from data management and data quality to creation, training and deployment of ML/DL models. For data collection, we used semi-structured interviews with data scientists, data analysts, AI research engineers, UX lead, ML engineers and technical leaders. The research approach as well as the roles and cases that were selected as the basis for this study are outlined in Figure~\ref{RolesCases}.

For analysis and coding of the empirical data, we adopted a thematic data analysis approach~\cite{maguire2017doing}. Following this approach, all cases were documented and we carefully reflected on our learnings and the implications of these. During analysis of our empirical findings, the interview transcripts were read carefully by the researchers to identify recurring elements and concepts, i.e. challenges experienced by practitioners in the cases we selected for this study~\cite{maxwell2012qualitative, eisenhardt1989building}. 

The details of the case studies, as well as a number of additional cases that were not selected for this particular paper, can be found in our previously published research~\cite{arpteg2018software, munappy2019data, lwakatare2019taxonomy}. In our previous research, we identified the challenges that practitioners experience when building ML/DL systems and we concluded that there is a significant need for future research on this topic. In this paper, and to advance our previous identification of challenges, we map the challenges to a set of strategic focus areas that we recognize in industry. Furthermore, we outline a research agenda for AI engineering research to help the research community structure and conceptualize the problem space. As recommended by~\cite{walsham1995interpretive}, the generalizations made based on case study research should be viewed as tendencies rather than predictions and as insights valuable for contexts with similar characteristics.  With the opportunity to work closely with more than a dozen large CPS
and SaaS companies, we believe that the insights we provide on the challenges these companies experience when building ML/DL systems will be valuable also outside the specific context of these companies. In addition, and as the main contribution of this paper, we believe that the research agenda we present based on the key engineering challenges surrounding ML/DL systems will provide support and structure for the research community at large.

\section{The Challenge of AI Engineering}\label{AIEchallenge}
Engineering AI systems is often portrayed as the creation of a ML/DL model,  and deploying it.
In practice, however, the ML/DL model is only a small part of the overall system and significant additional functionality is required to ensure that the ML/DL model can operate in a reliable and predictable fashion with proper engineering of data pipelines, monitoring and logging, etc.~\cite{bernardi2019150, sculley2015hidden}. To capture these aspects of AI engineering we defined the Holistic DevOps (HoliDev) model~\cite{bosch2018takes} [4], where we distinguish between requirements-driven development, outcome-driven development (e.g. A/B testing) and AI-driven development.

\subsection{AI adoption in practice}\label{AIadoption}

The challenge of AI engineering is that the results of each of the aforementioned types of development end up in the same system and are subject to monitoring of their behaviour as well as continuous deployment. In industrial deployments that we have studied, also AI models are constantly improved, retrained and redeployed and consequently follow the same DevOps process as the other software components.

\begin{figure}
\includegraphics[width=0.5\textwidth]{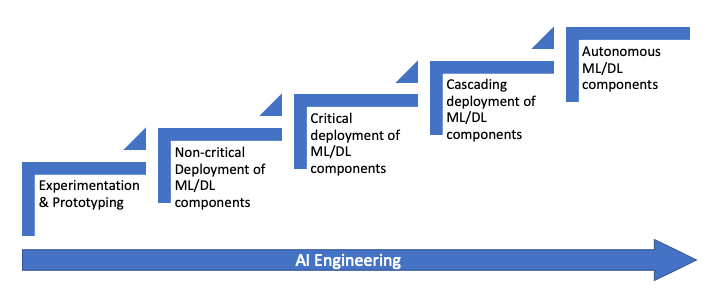}
\caption{The AI adoption evolution model}
\label{AI_adoption}
\vspace{-0.2 cm}
\end{figure}

In a transformation to AI-driven development, companies, over time, tend to develop more skills, capabilities and needs in the ML/DL space and consequently they evolve through several stages. In the AI Evolution model shown in figure~\ref{AI_adoption} we illustrate how companies, based on our research [5, 6], develop over time. The maturity of companies concerning AI evolves through five stages: 

\begin{itemize}
    \item {\bf Experimentation and prototyping}: This stage is purely exploratory and the results are not deployed in a production environment. Consequently, AI engineering challenges are not present in this stage.
    \item{\bf Non-critical deployment}: In this stage, a ML/DL model is deployed as part of a product or system in a non-critical capacity, meaning that if the model fails to perform, the overall product or system is still functional and delivers value to customers. 
    \item {\bf Critical deployment}: Once the confidence in the ML/DL models increases, key decision makers become sufficiently comfortable with deploying these models in a critical context, meaning that the product or system fails if the ML/DL model does not perform correctly.
    \item{\bf Cascading deployment}: With the increasing use of ML/DL models, the next step is to start to use the output of one model as the input for the next model in the chain. In this case, monitoring and ensuring correct functioning of the system becomes more difficult as the issues may be emergent, rather than directly associated with a specific ML/DL model.
    \item{\bf Autonomous ML/DL components}: In the final stage, ML/DL models monitor their own behaviour, automatically initiate retraining and are able to flag when the model observes that, despite retraining using the latest data, it does not provide acceptable accuracy.
\end{itemize}

Each step requires increased activities of “AI engineering” - a set of methods and tools that originated from software engineering in a system life cycle, and procedures, technologies and tools from data science and AI. While the first step, which is today state of the practice, typically covers the end-to-end ML development cycle (data acquisition, feature engineering, training and evaluation, and deployment), the next steps require the existing approaches from software engineering (e.g. system testing) as well as completely new methods that will need to become an integrated part of software and AI engineering (e.g. continuous training, or version management of code and data).

\subsection{AI engineering strategic focus}\label{AIfocus}

During our research, we have worked with a variety of companies and selected 16 cases for the purpose of this research. As part of our research, we have identified over 30 problems that are a concern in multiple cases that we have studied. We have presented some of these in earlier publications, specifically~\cite{arpteg2018software, lwakatare2019taxonomy, munappy2019data}, so we will not discuss each identified problem in detail. Instead, we provide an overview in figure~\ref{AI_adoption} and present a categorization of the identified problems in four strategic focus areas, relating to the typical phases of a ML project. These four areas are the following:

\begin{itemize}
    \item {\bf Data quality management}: One of the key challenges in successful AI projects is to establish data sets and streams that are of sufficient quality for training and inference. Specifically, data sets tend to be unbalanced, have a high degree of heterogeneity, lack labels, tend to drift over time, contain implicit dependencies and generally require vast amounts of pre-processing effort before they are usable.
    \item {\bf Design methods and processes}: Although creating an ML model is relatively easy, doing so at scale and in a repeatable fashion proves to be challenging. Specifically, managing a multitude of experiments, detecting and resolving implicit dependencies and feedback loops, inability of tracing data dependency, estimating effort, cultural differences between  developer roles, specifying desired outcome and tooling prove to be difficult to accomplish efficiently and effectively.
    \item {\bf Model performance}: The performance of ML/DL models depends on various factors, both for accuracy and for general quality attributes. Some of the specific problems that we have identified include a skew between training data and the data served during operation, lack of support for quality attributes, over-fitting of models and scaleable data pipelines for training and serving. 
    \item {\bf Deployment \& compliance}: Finally, one area that is highly underestimated is the deployment of models. Here, companies struggle with a multitude of problems, including monitoring and logging of models, testing of models, troubleshooting, resource limitations and significant amounts of glue code to get the system up and running. 
\end{itemize}

\begin{figure*}
\centering
\includegraphics[width=0.7\textwidth]{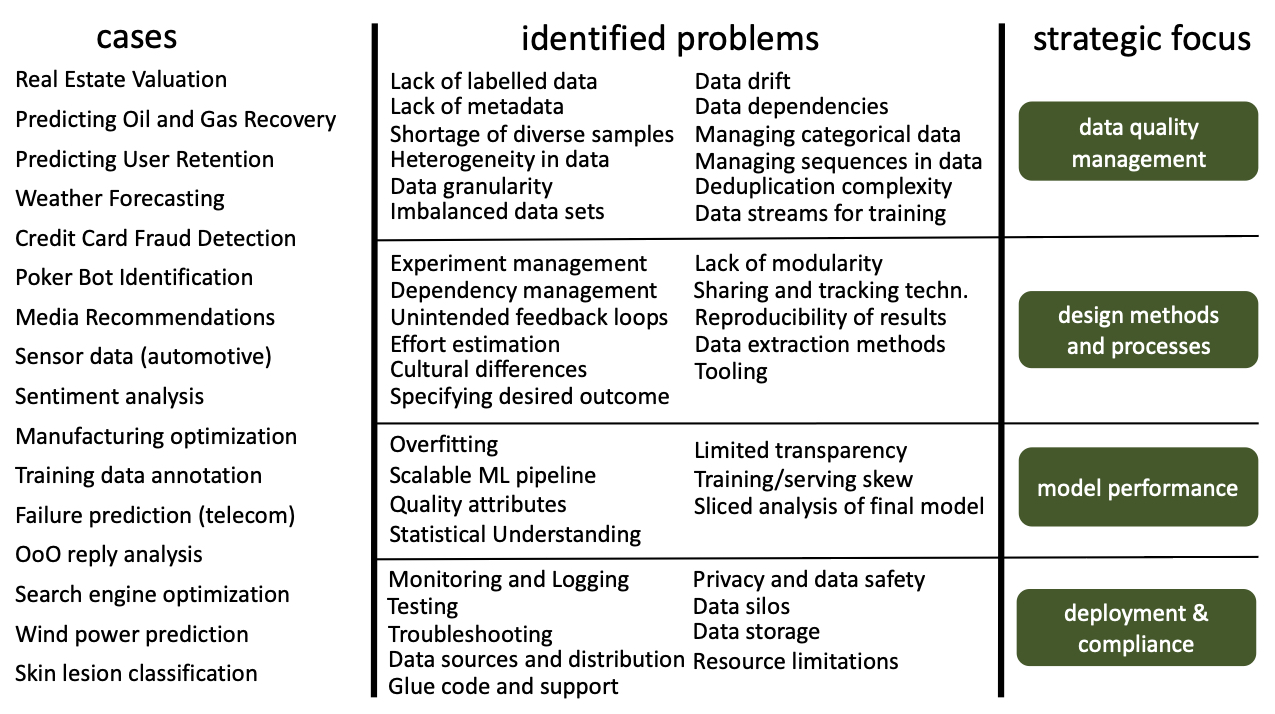}
\caption{Overview of cases and identified problems}
\label{CasesProblems}
\vspace{-0.2 cm}
\end{figure*}

\section{AI Engineering: A Research Agenda}\label{Agenda}
The subject of AI and the notion of engineering practices for building AI systems is a multi-faceted and complex problem. Consequently, few, if any, models exist that seek to create a structure and conceptualization of the problem space. Here we provide a structured view on the challenge of AI engineering and we provide a research agenda.
These challenges are organized into two main categories, i.e. \textit{generic AI engineering} and \textit{domain specific AI engineering}. Within generic AI engineering (AI Eng), we categorize the challenges into to three main areas, i.e. \textit{architecture}, \textit{development} and \textit{process}. For domain specific AI engineering (D AI Eng), we have identified one set of challenges for each domain that we have studied in the case study companies. 

\begin{figure*}
\includegraphics[width=\textwidth]{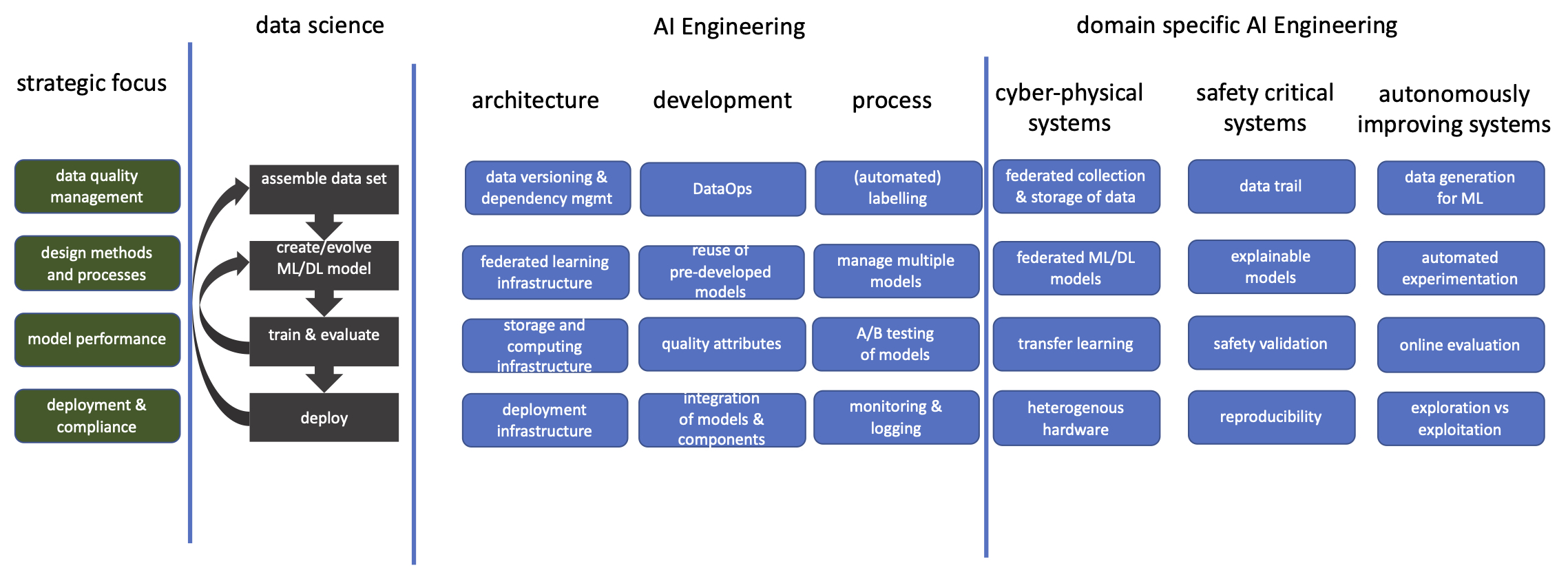}
\caption{Research agenda for AI engineering}
\label{ResearchAgenda}
\vspace{-0.2 cm}
\end{figure*}

As a second dimension, we follow the strategic focus areas that are related directly to the four main phases of a typical ML/DL project, i.e. data quality management (related to assembling data sets), design methods and processes (related to creating and evolving ML/DL models), model performance (related to training and evaluating) and finally deployment and conformance, related to the deploy phase. In figure~\ref{ResearchAgenda}, the model is presented graphically. In the remainder of the section, we discuss the key research challenges in more detail.

As the data science activities shown in figure~\ref{ResearchAgenda} are the regular AI/data science activities, we will discuss these only briefly:

\begin{itemize}
    \item {\bf Assemble data sets}: The first activity in virtually any ML/DL project is to assemble the data sets that can be used for training and evaluation and to evaluate these in order to understand the relevant features in the data.
    \item {\bf Create \& evolve ML/DL model}: After analysing the data sets, the next step is to experiment with different ML algorithms or DL models and to select the most promising one for further development. 
    \item {\bf Train \& evaluate}: Once the model has been developed, the next step is to train and validate the model using the data. 
    \item {\bf Deploy}: Once the model has been trained and is shown to have sufficient accuracy, recall and/or other relevant metrics, the model is deployed in a system where it typically is connected to one or more data streams for the purpose of inference.
\end{itemize}

The data science process above has many additional aspects and is typically conducted in an iterative manner. In figure~\ref{ResearchAgenda}, we show two of these iterations, i.e. between training and modeling and between deployment and the assembling of new data sets. However, as this paper is concerned with AI engineering and not with the specific data science aspects, we do not discuss these aspects in more detail.

\subsection{AI Engineering: Architecture}
In the context of AI engineering, architecture  is concerned with structuring the overall system and decomposing it into its main components. Constructing systems including ML/DL components require components and solutions not found in traditional systems and that need to address novel concerns. Below we describe the primary research challenges that we have identified in our research.

\begin{itemize}
    \item {\bf Data versioning \& dependency management}: The quality of the data used for training is absolutely central for achieving high performance of models. Especially in a DevOps environment, data generated by one version of the software is not necessarily compatible with the software generated by the subsequent version. Consequently versioning of data needs to be carefully managed. In addition, systems typically generate multiple streams of data that have dependencies on each other. As data pipelines tend to be less robust than software pipelines~\cite{munappy2019data}, it is important to provide solutions for the management of data quality. This can be concerned with simple checks for data being in range or even being present or more advanced checks to ensure that the average for a window of data stays constant over time or that the statistical distribution of the data remains similar. As ML/DL models are heavily data dependent, the data pipelines needed for feeding the models as well as the data generated by the models need to be set up. This can be particularly challenging when different types of data and different sources of data are used; in addition to questions of availability, accuracy, synchronisation and normalisation, significant problems related to security and privacy appear.
    \item {\bf Federated learning infrastructure}: Most of the cases that we studied concern systems where ML models are deployed in each instance of the system. Several approaches exist for managing training, evaluation and deployment in such contexts, but one central infrastructure component is the support for federated learning. As it often is infeasible to move all data to a central location for training a global model, solutions are needed for federated learning and the sharing of model parameters such as neural network weights as well as selected data sets that, for instance, represent cases not well handled by the central model. Federated learning requires an infrastructure to achieve the required quality attributes and to efficiently and securely share models and data.
    \item {\bf Storage and computing infrastructure}: Although many assume that all ML/DL deployments operate in the cloud, our interaction with industry shows that many companies build up internal storage and computing infrastructure because of legal constraints, cost or quality attributes. Developing these infrastructures, for example for the development of autonomous driving solutions, is a major engineering and research challenge. Typically collection and storing of data is organized centrally on the enterprise level, while development of AI solutions is distributed over several development teams. 
    \item {\bf Deployment infrastructure}: Independent of the use of centralized or federated learning approaches, models still need to be deployed in systems in the field. As most case study companies have adopted or plan to soon adopt DevOps, it is important for a deployment infrastructure to reliably deploy subsequent versions of models, measure their performance, raise warnings and initiate rollbacks in the case of anomalous behaviour. This infrastructure is by necessity of a distributed nature as it requires functionality both centrally as well as in each system that is part of the DevOps approach. Deployment of MD/DL models may require substantial change in the overall architecture of the system. 
\end{itemize}

\subsection{AI Engineering: Development}
Building and deploying successful ML/DL components and systems requires more than data science alone. In this section we focus on the development of systems including ML/DL components. This is important because also ML/DL models, in most cases that we have studied, are subject to the same DevOps activities as the other software in systems, meaning that models evolve, are retrained and deployed on continuous basis. Based on our case study research, we present the four primary research challenges concerning development in AI engineering below.

\begin{itemize}
    \item {\bf DataOps}: Although considered a buzzword by some, DataOps raises the concern of managing everything data with the same structured and systematic approach as that we manage software with in a traditional DevOps context. As typical companies ask their data scientists to spend north of 95\% of their time on cleaning, pre-processing and managing data, there is a significant opportunity to reduce this overhead by generating, distributing and storing data smarter in the development process. DataOps requires high levels of automation, which requires alignment and standardization in order to achieve continuous value delivery.
    \item {\bf Reuse of pre-developed models}: Most companies prefer to employ models developed by others or that have been developed earlier inside the company. However, reuse of existing ML/DL models is not trivial as the separation between the generic and specific parts of the model are not always easy to separate, in particular when the run-time context is different from that used in training phase.
    \item {\bf Quality attributes}: In data science, the key challenge is to achieve high accuracy, recall or other metrics directly related to the ML performance of the machine learning model. In an AI engineering context, however, several other quality attributes become relevant including the computation performance, in terms of the number of inferences per time unit the system can manage, the real-time properties, robustness of the system in case of data outside the scope of training set, etc. Ensuring satisfactory adherence to the quality requirements on the ML components in the system is a research challenge that is far from resolved.
    \item {\bf Integration of models \& components}: As we discussed earlier in the paper, ML/DL models need to be integrated with the remainder of the system containing regular software components. However, it is not always trivial to connect the data-driven ML/DL models with the computation-driven software components. Also, traditional testing and evaluation of the models must be integrated in such a way that software methods and data-science evaluation methods are combined seamlessly. Depending on the criticality of the ML/DL model for the overall performance of the system, the validation activities need to be more elaborate and strict. 
\end{itemize}

\subsection{AI Engineering: Process}
Although the notion of process has gone out of vogue with the emergence of agile, it is hard to argue that no process is required to align the efforts of large groups of people without prohibitively high coordination cost. The context of AI engineering is no different, but there are surprisingly few design methods, processes and approaches available for the development and evolution of ML/DL models. Experienced data scientists do not need these, but with the rapidly growing need for AI engineers, many less experienced data scientists and software engineers are asked to build these models. These professionals would very much benefit from more methodological and process support. We have identified four main process related challenges that require significant research efforts to resolve in a constructive and efficient way. Below we describe each of these in more detail.

\begin{itemize}
    \item {\bf Automated labelling}: As the data sets that a company starts with are limited sources for training and validation, ideally we want to collect the data sets for training evolving models during operation in deployment. Although it is easy to collect the input data, the labels used in supervised learning are often much harder to add. Consequently, we need solutions for, preferably, automated labelling of data so that we have a constant stream of recent data for training and validation purposes during evolution.
    \item {\bf Manage multiple models}: The first concern that often surfaces in teams working on ML/DL models is that it is difficult to keep track of all the models that are being considered during the development phase. We discussed parts of this challenge in [2].
    \item {\bf A/B testing of models}: During evolution, the improved model is deployed for operation. However, experience shows that models that perform better in training do not necessarily perform better in operations. Consequently, we need solutions, often variants of A/B testing, to ensure that the new model also performs better in deployment.
    \item {\bf Monitoring \& logging}: Once the model is deployed and used in operation, it is important to monitor its performance and to log events specific to the performance of the model. As ML/DL models tend to lack on the explainability front, the monitoring and logging is required to build confidence in the accuracy of the models and to detect situations where the performance of a model starts to deteriorate or is insufficient from the start.
\end{itemize}

\subsection{Domain-specific AI Eng: Cyber physical systems}
In the remainder of this section, we present the unique research topics for three application domains in which ML/DL technologies are being deployed, i.e. cyber physical systems, safety critical systems and autonomously improving systems. Our research shows that each domain brings with it a set of unique activities and research challenges associated with AI engineering topics.

Although the recent emergence of ML/DL models in industry started in the online SaaS world, this has been rapidly followed by increasing interest in the software-intensive embedded systems industry. The main difference with cloud based deployments is that the ML/DL models are deployed in embedded systems out in the field such as base stations, cars, radars, sensors and the like. 

Cyber physical systems are often organized around three computing platforms, i.e. the edge device where the data for ML/DL is collected, an on-premise server of some kind and the infrastructure in the cloud. Each of these platforms has its own characteristics in terms of real-time performance, security and privacy, computational and storage resources, communications cost, etc.

The consequence is that data management, training, validation and inference associated with ML/DL models has a tendency to become federated as it requires these three computing platforms as most capabilities that customers care about will cross-cut all three platforms. This leads to a set of unique research challenges for this domain that we discuss below.

\begin{itemize}
    \item {\bf Federated/distributed storage of data}: Parallel to the model, the data used for training and inference needs to be managed in a distributed and federated fashion. Local storage on device instances minimizes communication cost, but tends to increase the bill-of-materials for each device and these architectural drivers need to be managed. 
    \item {\bf Federated/distributed model creation}: Due to the presence of multiple computing platforms, the architect or data scientist needs to distribute the ML/DL model over these computing platforms, resulting in a federated model. This is an open research area related to the system and data lifecycles, performance, availability, security, computation, etc. 
    \item {\bf Transfer learning}: Especially for companies that have thousands or millions of devices deployed in the field, the challenge is the balancing between centralized and decentralized learning. The most promising approach is to distribute centrally trained models and to allow each individual device to apply its local learnings to the centrally trained model using transfer learning approaches. However, more research is needed.
    \item {\bf Deploy on heterogeneous hardware}: Finally, because of both cost and computational efficiency, embedded systems often use dedicated hardware solutions such as ASICs and FPGAs. Additionally, MD/DL models require huge amounts of parallel computation, both during training and implementation, realised in e.g. GPUs. These execution platforms use different development environments, programming languages, and execution paradigms. Embedded systems tend to have constraints on computational and storage resources as well as power consumption. Deploying ML/DL models on these types of hardware frequently requires engineering effort from the team as there are no generic solutions available.
\end{itemize}

One challenge that is not yet one of the primary ones but that has appeared on the horizon is mass-customization of ML/DL models. As some CPS companies have many instances of their products in the field, the ML/DL models deployed in these instances should, ideally, adjust their behaviour to the specifics of the users using the instance, i.e. mass-customization. However, there are few solutions available for combining both continuous deployment of centrally trained models with the customization of each product instance.

\subsection{Domain-specific AI Eng: Safety-critical systems}
A special class of cyber physical systems are safety-critical systems, i.e. those systems whose failure or malfunction may result in significant bodily, environmental or financial harm. The community struggles with balancing two forces. On the one hand, we seek to avoid harm by taking conservative approaches and introducing new technologies only after careful evaluation. On the other hand, the slow introduction of new technologies may easily cause harm in that the new technologies can help avoid safety issues that were not possible to avoid with conventional technologies only.

One of these new technologies is, of course, ML/DL. In the automotive industry, among others, the use of ML/DL allows for advanced driver support functions as well as fully autonomous driving. The open challenge is establishing the safety of these systems. In our research, we have defined the four primary research challenges specific for safety-critical AI-based systems.

\begin{itemize}
    \item {\bf Data trail}: One of the key challenges in safety critical systems is that the collection of safety-related evidence before the deployment of systems and the creation of a data trail during operations in order to ensure safe operation of the system. In the context of ML/DL models, this requires maintaining a clear trail of the data that was used for training as well as the inferences that the model provided during operation. Little research exists that addresses this challenge for AI components and consequently this is a significant research challenge.
    \item {\bf Explainable models}: As it is virtually impossible to test a system to safety, the community often uses various approaches to certify systems. This is performed by assessors who need to understand the functionality of the system. This requires that ML/DL models are explainable, which today is unsolvable or at least a non-trivial problem for most models. 
    \item {\bf Validation of safety-critical systems}: The basic enabler for deployment of ML/DL models in safety critical systems is the validation of these systems. Validation concerns both the correct behaviour in situations where application should act, but we also need to show that the system will not engage in situations where it is not necessary or even dangerous to do so. Validation of safety-critical systems starts from requirements of justifiable prediction and of deterministic system behavior, while ML/DL solutions are based on statistical models, so in principle non-deterministic behavior. In practice, the ML/DL models can be more accurate and reliable, but justification of these models requires new approaches, methods, and standards in the validation process. 
    \item {\bf Reproducibility}: For a variety of factors, a ML/DL model may end up looking different when it is given a different seed, order of training data, infrastructure it is deployed on, etc. Especially for safety critical systems, it is critical that we can reproduce the model in a predictable manner, independent of the aforementioned factors.
\end{itemize}

\subsection{Domain-specific AI Eng: Autonomously improving systems}
There is an emerging category of systems that uses ML/DL models with the intent of continuously improving the performance of the system autonomously. In practice, there are humans involved in the improvement of the system, but the system employs mechanisms for experimentation and improvement that do not require human involvement. 

The primary way for systems to achieve this is through the use of ML/DL models that analyse the data, train using it and then provide interference. This requires forms of automated experimentation where the system itself generates alternatives and, for example, deploys these alternatives in A/B testing or similar contexts and measures the impact of these changes. There are four research challenges challenges that need to be addressed for autonomously improving systems:

\begin{itemize}
    \item {\bf Data generation for machine learning}: Traditional ML/DL model development requires data scientists to spend significant amounts of time to convert available data sets that often are intended for human consumption into data sets that are usable for machine learning. In autonomously improving systems, the data that is generated by the system needs to be machine interpretable without any human help. How to accomplish this, though, is an open research question.
    \item {\bf Automated experimentation}: Although the notion of automated experimentation is conceptually easy to understand, actually realizing systems that can operate in this fashion is largely an open research challenge where little work is available.
    \item {\bf Online evaluation}: As autonomously improving systems generate alternatives for evaluation at run-time, these alternatives need to be deployed and evaluated during the regular operation of the system. This requires solutions for dynamically adjusting the behavior of the system to select, for a small percentage of the cases, the new alternative for evaluation as well as to keep track of statistical validity of the test results associated with each alternative.
    \item {\bf Exploration vs exploitation}: In autonomously improving systems, the system autonomously experiments with different responses to the environment in which it operates with the intent of improving its performance. The challenge is that some or even many of these experiments will result in worse performance than the current default response. This is referred to as {\it regret} in optimization theory. The challenge is that it is impossible to find better ways of doing things without trying out these new ways, but especially in embedded systems there is a limit to how poor the alternative can be. This means that we need research to help assess the worst case outcomes for each experiment with the intent of balancing the cost of exploration with the cost too much exploitation.
\end{itemize}

\subsection{Other domain specific systems}
We described the domain specific research challenges for building ML/DL systems for specific types of systems. There of course are other domains that likely have specific research challenges as well. These challenges might be the same as for non AI-based systems, but new methods must be developed to meet these challenges (for example develop new methods to ensure system reliability, availability, security, reusability, or other non-functional properties). However, in many cases introducing ML/DL solutions cause new challenges such as quality of data, real-time data access, increase in efforts in the development life cycle as well as challenges in combination of security, functionality and privacy, etc.

\section{Conclusion}~\label{Conclusion}
Artificial intelligence, and specifically machine- and deep-learning, has, over the last decade, proven to have the potential to deliver enormous value to industry and society. This has resulted in most companies experimenting and prototyping with a host of AI initiatives. Unfortunately, our research~\cite{arpteg2018software, lwakatare2019taxonomy, munappy2019data} shows that the transition from prototype to industry-strength, production-quality deployment of ML models proves to be very challenging for many companies. The engineering challenges surrounding this prove to be significant~\cite{sculley2015hidden}, even if many researchers and companies fail to recognize this.

To the best of our knowledge, no papers exist that provide a systematic overview of the research challenges associated with the emerging field of AI engineering. In this paper, we provide a conceptualization of the typical evolution patterns that companies experience when adopting ML, present an overview of the problems that companies experience based on well over a dozen cases that we have studied and we provide a research agenda that has been derived from the research that we have conducted to date and that needs to be addressed by the research community at large. The goal of this research agenda is to provide inspiration for the software engineering research community to start addressing the AI engineering challenges.

AI and ML have the potential to greatly benefit industry and society at large. For us to capture the value, however, we need to be able to engineer solutions that deliver production-quality deployments. This requires research to address the AI engineering challenges that we present in this paper. In future work, we aim to address several of these research challenges in our research and our collaboration with industry. In particular collaboration with industry in real industrial settings is crucial since ML methods build upon empirical methods and directly depend on the amount and types of data. For this reason, we frequently organize events in the Nordics and at international conferences to create awareness for the identified challenges and to encourage other researchers to join us in addressing these.

\section*{Acknowledgment}
The research in this paper has been supported by Software Center, the Chalmers Artificial Intelligence Research Center (CHAIR) and Vinnova.

\bibliographystyle{IEEEtran}
\bibliography{AIEngineeringResearchAgenda.bib}

\end{document}